\documentclass{article}
\usepackage{ijcai22}

\usepackage{times}

\usepackage{soul}
\usepackage[utf8]{inputenc}
\usepackage[small]{caption}
\usepackage{graphicx}
\usepackage{amsmath}
\usepackage{booktabs}
\usepackage{tikz}
\usepackage{xspace}
\usepackage{algorithm2e}

\usepackage{mathtools}
\usepackage{amsmath}
\usepackage{amsthm}
\usepackage{amsbsy}
\usepackage{amssymb}
\usepackage{amsfonts}
\usepackage{nicefrac}
\usepackage{bm}

\usepackage{tikz}
\usetikzlibrary{decorations.markings}
\usetikzlibrary{arrows}
\usetikzlibrary{shapes}
\usetikzlibrary{positioning}
\usepackage{pgfplots}
\pgfplotsset{compat=1.10}

\usepackage{etoolbox}
\usetikzlibrary{shapes.geometric,arrows,fit,matrix,positioning}
\tikzset
{
    treenode/.style = {circle, draw=black, align=center, minimum size=1.1cm},
}

\newtheorem{example}{Example}

\newtheorem{definition}{Definition}
\newtheorem{proposition}{Proposition}

\renewcommand{\vec}[1]{\bm{#1}}
\renewcommand{\L}{\mathcal{L}}

\def\sdd{{\tt SDD}}
\def\obdd{{\tt OBDD}}
\def\adt{{\tt ADT}}
\def\dt{{\tt DT}}
\def\rf{{\tt RF}}

\def\nc{\mathbf{\neg C}}
\def\bac{\mathbf{\wedge BC}}
\def\boc{\mathbf{\vee BC}}

\title{Rectifying Mono-Label Boolean Classifiers}

\author{
Sylvie Coste-Marquis$^1$\and
Pierre Marquis$^{1,2}$ 
\affiliations
$^1$Univ. Artois, CNRS, CRIL\\
$^2$Institut Universitaire de France\\
\emails
\{coste, marquis\}@cril.fr
}

\begin{document}

\maketitle

\begin{abstract}
We elaborate on the notion of \emph{rectification of a Boolean classifier} $\Sigma$ 
introduced in \cite{DBLP:conf/ijcai/Coste-MarquisM21}. Given $\Sigma$ and some background 
knowledge $T$, postulates characterizing the way $\Sigma$ must be changed into a new classifier $\Sigma \star T$ 
that complies with $T$ were presented. 
We focus here on the specific case of \emph{mono-label} Boolean classifiers, i.e., there is a single target concept and any instance
is classified either as positive (an element of the concept), or as negative (an element of the complementary concept).
In this specific case, our main contribution is twofold:
(1) we show that there is a \emph{unique} rectification operator $\star$ satisfying the postulates,  and
(2) when $\Sigma$ and $T$ are Boolean circuits, we show how a classification circuit equivalent 
to $\Sigma \star T$ can be computed in \emph{time linear} in the size of $\Sigma$ 
and $T$; when $\Sigma$ and $T$ are decision trees, a decision tree 
equivalent to $\Sigma \star T$ can be computed in \emph{time polynomial}  in the size of $\Sigma$ 
and $T$.
\end{abstract}

\section{Introduction}

Much work has been devoted for the past few years to eXplainable AI, in the objective of making ML models less opaque, see e.g., 
\cite{Adadi.IEEE.2018,Miller19,Samek.BOOK.2019,GuidottiMRTGP19,DBLP:conf/ijcai/0002C20,Molnar.BOOK.2020,DBLP:journals/natmi/LundbergECDPNKH20,DBLP:journals/corr/abs-2103-11251}). 
This went typically through the definition of a number of explanation and/or verification issues for various ML models, and the development 
and the evaluation of algorithms for addressing those issues. 

Verifying a model requires to be able to test whether the
predictions made by the model are correct or not, and this often asks for leveraging the skills of an expert. Whenever the prediction made is viewed as 
incorrect or, more generally, when it conflicts with the expert knowledge, a more challenging issue is to figure 
out \emph{how the ML model should be modified} to ensure that the prediction made will be correct afterwards, and that the predictor 
will comply with the expert knowledge.

To make things concrete, let us consider the following credit scoring scenario. Alice, a bank employee, receives Bob, a customer who wants to obtain a loan. The bank management provides Alice with an AI algorithm (a predictor) to help her decide which issue to give to any loan application. Alice uses this algorithm and it recommends against granting Bob the requested loan. Alice is very surprised by the result provided by the algorithm, since she is experienced and has already granted loans to clients of the bank with precisely the same profile as Bob’s, i.e., a client with low incomes but who has reimbursed a previous loan and has no debts. Alice’s expertise led her not to follow the recommendation of the AI algorithm and to grant Bob the loan requested. However, Alice would like to do more to avoid that the problem encountered arises again with another client having an identical profile. She wonders what could be done to this end.

The research question tackled in our recent work \cite{DBLP:conf/ijcai/Coste-MarquisM21} is relevant to Alice’s concern. 
In this work, we introduced a change operation, called \emph{rectification}, that is suited 
to multi-label Boolean classifiers. Given a set $X = \{x_1,\cdots, x_n\}$ (its elements are Boolean features) 
and a set $Y = \{y_1,\cdots, y_m\}$, that is disjoint with $X$ (its elements are labels, denoting classes / concepts),
$\vec X$ is the set $\{0,1\}^n$ of all vectors over $\{0, 1\}$ of size $n$, and 
$\vec Y$ is the set $\{0,1\}^m$ of all vectors over $\{0, 1\}$ of size $m$.
Then, a multi-label Boolean classifier simply is a mapping $f$ from $\vec X$ to $\vec Y$, associating with
each input instance (a vector $\vec x \in \vec X$ of $n$ Boolean values assigned to the elements of $X$)
a vector $\vec y \in \vec Y$ of $m$ Boolean values assigned to the elements of $Y$. Whenever an instance
$\vec x = (x_1, \ldots, x_n)$  is associated by the classifier with
$\vec y = (y_1, \ldots, y_m)$ such that $y_i$ ($i \in [m]$) is equal to $1$ (resp. $0$),
one considers that $\vec x$ belongs to the class $y_i$ (resp. does not belong to this class).

For instance, considering the previous credit scoring scenario, one may assume the following sets of Boolean features $X = \{x_1, x_2, x_3\}$ and labels $Y = \{y\}$,
associated with the following semantics:
\begin{itemize}
%\item $x_1 = 1$: has low income
%\item $x_2 = 1$: has reimbursed a previous loan
%\item $x_3 = 1$: has debts
%\item $y = 1$: to grant the loan
\item $x_1 = 1$: has low income
\item $x_2 = 1$: has reimbursed a previous loan
\item $x_3 = 1$: has debts
\item $y = 1$: to grant the loan
\end{itemize}
Bob is viewed as the instance $\vec x = (x_1 = 1, x_2 = 1, x_3 = 0)$, and if $f$ denotes the predictor used by Alice, we have $f(\vec x) = 0$, meaning
that the predictor suggests not to grant the loan.

A multi-label Boolean classifier $f$ can be represented by a Boolean circuit $\Sigma$ over a set of variables $\mathit{PS}$ 
such that $X \cup Y \subseteq \mathit{PS}$. In such a circuit $\Sigma$, called a \emph{classification circuit}, features and labels are both represented by propositional variables despite the fact that they correspond to distinct notions. 
Some pieces of knowledge supposed to be more reliable than the classification circuit $\Sigma$ are also considered.
They are represented as well by a Boolean circuit $T$ over $\mathit{PS}$.
The purpose of rectifying the classification circuit $\Sigma$ by $T$ is to modify $\Sigma$ so that (1) the constraints imposed by
$T$ on the facts about $Y$ that must hold under each $\vec x$ are respected, and (2) the resulting circuit noted $\Sigma \star T$ is still a
classification circuit. A minimal change condition is taken into account; it states that the way $\vec x$ was classified by $\Sigma$ 
must not be modified by the rectification process when the constraints imposed by $T$ on the facts that hold under $\vec x$ are already satisfied.
In the general case when $Y$ is unconstrained, several classification circuits $\Sigma \star T$ can be found that satisfy (1) and the minimal change condition.
Stated otherwise, several rectification operators $\star$ can be defined.
 
In  \cite{DBLP:conf/ijcai/Coste-MarquisM21},  it was shown that a rectification operation amounts to a specific form of \emph{belief change} \cite{citeulike:4115507}.  
A logical characterization of classification circuits has been pointed out and a
number of postulates that rectification operators should satisfy have been presented. We also exhibited some operators
from the rectification family, and studied the standard belief change postulates in order to determine those that are satisfied by
every rectification operator satisfies, and those that are not. Especially, we proved that the families of rectification operators 
and those of ``standard'' belief change operators, namely revision operators / update operators \cite{KM,KMupdate}, are disjoint. 

In this paper, we focus on the specific case of \emph{mono-label Boolean classifiers}, i.e., there are only two classes,
the target concept and the complementary one, so that every instance is either positive (i.e., it belongs to the target concept) or negative
(i.e., it does not belong to it). This is ensured by considering that $Y = \{y\}$ is a singleton (as it it the case for the credit scoring scenario). 
Under this assumption, we present two contributions. On the one hand, provided that the Boolean classifier is represented by a circuit $\Sigma$ involving only variables $X$ for the representation of instances and the concept variable $y$, we show that there exists a \emph{unique rectification operator}, noted $\star$, 
thus providing a full characterization of rectification operators in this restricted case. Since it is unique, $\star$ coincides with the operators 
presented in \cite{DBLP:conf/ijcai/Coste-MarquisM21} when $Y$ is a singleton. On the other hand, we show how a classification circuit equivalent to $\Sigma \star T$ can be computed in \emph{time linear} in the size of $\Sigma$ and $T$, where the change formula $T$ is given as a Boolean circuit. This result fully 
contrasts with the representations of rectified classifiers presented in \cite{DBLP:conf/ijcai/Coste-MarquisM21}, which are of size 
exponential in the size of $X$. In addition, the specific cases when $\Sigma$ and $T$ are \sdd\ circuits \cite{Darwiche11}, \obdd\ circuits \cite{Bryant86}, and 
(possibly affine) decision trees \cite{Wegener00,Koricheetal13} are analyzed.

The rest of the paper is organized as follows. Some formal preliminaries are provided in Section \ref{sec:prelim}. Our characterization theorem is presented
in Section \ref{sec:carac}. Our representation result is given in Section \ref{sec:representing}. Section \ref{sec:conclusion} concludes the paper.

\section{Preliminaries}\label{sec:prelim}

Before presenting the key definitions of the rectification setting pointed out in \cite{DBLP:conf/ijcai/Coste-MarquisM21}, we first need
to recall a couple of notions about propositional representations.

A \emph{Boolean circuit} $\Phi$ over a set $\mathit{PS}$ of propositional variables is a DAG where internal nodes are labelled by usual 
connectives, $\neg$, $\vee$, $\wedge$, but may also correspond to decision nodes over variables from $\mathit{PS}$,
and leaves are labelled by variables from $\mathit{PS}$ or by Boolean constants $\top$ - {\it verum} - and $\bot$ - {\it falsum} (also 
denoted $1$ and $0$, respectively).  The size $|\Phi|$ of $\Phi$ is the number of arcs in it. 
$\mathit{Var}(\Phi)$ the subset of variables of $\mathit{PS}$ occurring in $\Phi$. $\L$ is the set of all Boolean circuits over $\mathit{PS}$.
When $X \subseteq \mathit{PS}$, $\L_X$ denotes the subset of $\L$ consisting of Boolean circuits over $X$.

For any node $N$ of $\Phi$, let $\Phi_N$ be the subcircuit of $\Phi$ rooted at node $N$, i.e., the subgraph of $\Phi$ 
that consists of all the nodes and arcs that can be reached from $N$.
Whenever $N$ is a decision node labelled by variable $x \in X$ in a Boolean circuit $\Phi$, the subcircuit $\Phi_N$ of $\Phi$ given by

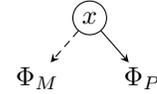
\begin{center}
\begin{tikzpicture}
\node[circle, draw, inner sep=2] (x) at (0,0) {$x$};
\node[inner sep = 1.5] (u) at (-0.7,-0.8) {$\Phi_M$};
\node[inner sep = 1.5] (v) at (+0.7,-0.8) {$\Phi_P$};
\draw[-stealth, densely dashed] (x) -- (u);
\draw[-stealth] (x) -- (v);
\end{tikzpicture}
\end{center}

is viewed as a short for the Boolean circuit

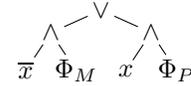
\begin{center}
\begin{tikzpicture}
\node[inner sep = 1.5] (or) at (0,0) {$\lor$};
\node[inner sep = 1.5] (a1) at (-1+1/3,-0.3) {$\land$};
\node[inner sep = 1.5] (a2) at (2/3,-0.3) {$\land$};
\node[inner sep = 1.5] (nx) at (-1,-0.8) {$\overline{x}$};
\node[inner sep = 1.5] (x) at (1/3,-0.8) {$x$};
\node[inner sep = 1.5] (u) at (-1+2/3,-0.8) {$\Phi_M$};
\node[inner sep = 1.5] (v) at (+1,-0.8) {$\Phi_P$};
\draw (a2) -- (or) -- (a1);
\draw (u) -- (a1) -- (nx);
\draw (v) -- (a2) -- (x);
\end{tikzpicture} % (see Definition 2.6 and Figure 2 in~\cite{DarwicheM02}).
\end{center}

A \emph{formula} over $\mathit{PS}$  simply is a Boolean circuit over $\mathit{PS}$  where the underlying DAG is a tree.
A \emph{literal} is a propositional variable of $\mathit{PS}$, possibly negated, 
or a Boolean constant. 
Any propositional variable $x$ is called a \emph{positive literal}, 
and the negation of $x$, denoted $\neg x$ or $\overline x$, is called a \emph{negative literal}.
%If $\ell$ is a literal $x$ (resp. $\neg x$), then its complementary literal $\sim \ell$ is $\neg x$ (resp. $x$).
For any subset $X$ of $\mathit{PS}$, $L_{X}$ denotes the set of literals based on the variables of $X$.
A \emph{term} is a conjunction of literals, and a \emph{clause} is a disjunction of literals.
A \emph{canonical term} over $X$ is a consistent term into which every variable of $X$ occurs (as such, or negated).
In the following, every instance $\vec x \in \vec X$ is also viewed as a canonical term over $X$, still noted $\vec x$, such that for every $i \in [n]$, $x_i$
is a literal of this term if the $i^{th}$ coordinate of $\vec x$ is $1$ and $\overline{x_i}$ is a literal of this term otherwise.

Given a set of variables $V \subseteq \mathit{PS}$, an \emph{interpretation over $V$} is a mapping $\omega$ from $V$ to $\mathbb{B} = \{0, 1\}$.
Every interpretation over $V$ corresponds to a unique canonical term over $V$, and vice versa.
When a total ordering $<$ over $\mathit{PS}$ is provided, interpretations can be represented by bit vectors or by words. For instance, if $V = \{v_1, v_2\}$ such that
$v_1 < v_2$, then the mapping $\omega$ such that $\omega(v_1) = 0$ and $\omega(v_2) = 1$ can be represented by the vector $(0, 1) \in \{0, 1\}^2$ or equivalently by
the word $01$.
Boolean circuits are interpreted in a classical way.
For a Boolean circuit $\Phi \in \L_V$ and an interpretation over any superset of $V = \mathit{Var}(\Phi)$, 
we use $\omega \models \Phi$ to denote the fact that $\omega$ if a \emph{model} of $\Phi$  
according to the semantics of propositional logic. That is, assigning the variables of $\Phi$
to truth values as specified by $\omega$ makes $\Phi$ true.
By $[\Phi]$ we denote the \emph{set} of models of $\Phi$ over $\mathit{Var}(\Phi)$.
$\Phi$ is \emph{inconsistent} if $[\Phi] = \emptyset$, and $\Phi$ is \emph{consistent} otherwise.  
A Boolean circuit $\Psi \in \L_V$ is a \emph{logical consequence} of a Boolean circuit $\Phi \in \L_V$ 
(denoted $\Phi \models \Psi$) if $\Phi \wedge \neg \Psi$ is inconsistent.
$\Phi$ and $\Psi$ are  \emph{logically equivalent} 
(denoted $\Phi \equiv \Psi$) if they have the same models over $\mathit{Var}(\Phi) \cup \mathit{Var}(\Psi)$.

Given a Boolean circuit $\Phi \in \L$ and a consistent term $\gamma$ over $\mathit{PS}$, the \emph{conditioning} of $\Phi$ 
by $\gamma$ is the Boolean circuit from $\L$, noted $\Phi(\gamma)$, obtained by replacing in $\Phi$ every occurrence of a variable $v \in \mathit{Var}(\gamma)$ by
$\top$ if $v$ is a positive literal of $\gamma$ and by $\bot$ if $\overline{v}$ is a negative literal of $\gamma$.

When $V$ is a subset of $\mathit{PS}$, a Boolean circuit $\Phi \in \L$ 
is said to be {\em independent} of $V$ if there is a Boolean circuit $\Psi \in \L$
logically equivalent to $\Phi$ such that $\mathit{Var}(\Psi) \cap V = \emptyset$.  
The {\em forgetting} of \(V\) in $\Phi$, 
denoted \(\exists V . \Phi\), is  (up to logical equivalence) the {\em strongest logical consequence} 
of $\Phi$ that is independent of $V$ (see e.g., \cite{LLM03}).
The \emph{projection} of $\Sigma$ onto $V$ is the forgetting of $\overline V$ in $\Sigma$, where $\overline V$ denotes
the set $\mathit{PS} \setminus V$.
$\exists V . \Phi$ can be characterized as follows:
\begin{itemize}
\item $\exists \emptyset . \Phi \equiv \Phi$,
\item $\exists \{v\} . \Phi \equiv (\Phi(\overline{v}) \vee (\Phi(v))$,
\item $\exists  (V' \cup \{v\}) . \Phi \equiv \exists V' . (\exists \{v\} . \Phi)$.
\end{itemize}

Let $\vec x \in \vec X$. A Boolean circuit $\Phi$ over $\L$ is said to {\em classify} 
$\vec x$ as $\vec v$ if and only if the Boolean circuit $\Phi(\vec x)$ has a unique model over $V = \mathit{PS} \setminus X$, given by $\vec v$. 
$\Phi$ has \emph{the $XY$-classification property} if and only if $Y \subseteq \mathit{PS} \setminus X$ and $\Phi$ classifies every $\vec x \in \vec X$. 
In that case, one also says that $\Phi$ is \emph{a classification circuit}.
When $\mathit{PS} \setminus X = \{y\}$ is a singleton, a Boolean circuit $\Phi$ is said to classify $\vec x$ as a \emph{positive instance} if $\Phi(\vec x) \equiv y$, 
as a \emph{negative instance} if $\Phi(\vec x) \equiv \overline{y}$, and $\Phi$ does not classify $\vec x$ in the remaining case.

The last notion to be made precise before defining rectification operators is the notion of \emph{fact compliance}.
A Boolean circuit $\Sigma \in \L$ is {\em fact-compliant} with a Boolean circuit $T \in \L$ on an instance 
$\vec x$ if and only if $\Sigma(\vec x) \models F(T, \vec x)$ where
\begin{tabular}{ll}
$F(T, \vec x)$ & = $\top$ if $T(\vec x)$ is inconsistent,\\
& $= \bigwedge_{\ell \in L_{\mathit{Y}} \mbox { s.t. } T(\vec x) \models \ell} \ell$ otherwise.
\end{tabular}

When $T(\vec x)$ is consistent, $F(T, \vec x)$ is the conjunction of all the facts (literals) about $Y$ that hold in $T(\vec x)$.
Accordingly, for every $T$ and every $\vec x$, we have $T(\vec x) \models F(T, \vec x)$.

\begin{example}\label{ex:example1}
Let $X = \{x_1, x_2\}$ and $Y = \{y_1, y_2\}$. Let $\Sigma$ be the Boolean circuit over $X \cup Y$ given by 
Figure \ref{fig:example1}. $\Sigma$ is logically equivalent to $(x_1 \Leftrightarrow y_1) \wedge (x_2 \Leftrightarrow y_2)$. 
The instance $(1, 1)$ corresponds to the canonical term $x_1 \wedge x_2$. Similarly,
the instance $(1, 0)$ corresponds to the canonical term $x_1 \wedge \overline{x_2}$. One can easily verify that
$\Sigma((1,1)) \equiv y_1 \wedge y_2$, $\Sigma((1,0)) \equiv y_1 \wedge \overline{y_2}$, $\Sigma((0,1)) \equiv \overline{y_1} \wedge y_2$, and
$\Sigma((0,0)) \equiv \overline{y_1} \wedge \overline{y_2}$. Thus, $\Sigma$ classifies every instance $\vec x$, and
as such, $\Sigma$ is a classification circuit.

\begin{figure}
\centering
    \scalebox{0.7}{
          \begin{tikzpicture}[scale=0.475, roundnode/.style={circle, draw=gray!60, fill=gray!5, thick, minimum size=5mm},
          squarednode/.style={rectangle, draw=blue!60, fill=blue!5, thick, minimum size=3mm}]           
            \node(root) at (7,7){$\wedge$};
            \node[roundnode](n1) at (3,5){$x_1$};
            \node[roundnode](n2) at (12,5){$y_2$};
            \node(n11) at (1,3){$\neg$};
            \node(n111) at (1,1){$y_1$};            
            \node[roundnode](n12) at (5,3){$y_1$};
            \node[squarednode](n121) at (4,1){$0$};
            \node[squarednode](n122) at (6,1){$1$};
            \node(n21) at (9,3){$\neg$};
            \node(n211) at (9,1){$x_2$};       
            \node(n22) at (15,3){$x_2$};            
            \draw(root) -- (n1);
            \draw(root) -- (n2);
            \draw[dashed] (n1) -- (n11);
            \draw(n1) -- (n12); 
            \draw[dashed] (n2) -- (n21);
            \draw(n2) -- (n22);    
            \draw[dashed] (n12) -- (n121);
            \draw(n12) -- (n122); 
            \draw(n11) -- (n111);      
            \draw(n21) -- (n211);                     
            \end{tikzpicture}
     }   
\caption{\label{fig:example1} A classification circuit.}
\end{figure}
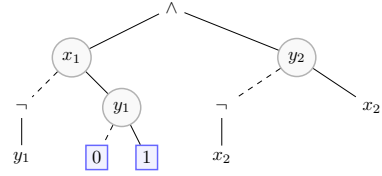

Now, let $T = ((x_1 \wedge x_2) \Rightarrow (y_1 \wedge y_2)) \wedge ((x_1 \wedge \overline{x_2}) \Rightarrow (y_1 \vee y_2)) \wedge ((\overline{x_1} \wedge x_2) \Rightarrow \overline{y_2}) \wedge (x_1 \vee x_2)$. $T$ is a formula and it does not have the $XY$-classification property. Indeed, though $T$ classifies $(1, 1)$ (as $(1, 1)$), it does not classify any of the other instances: $T((1,0))$ has three models over $Y$, $T((0, 1))$ has two models over $Y$, and $T((0, 0))$ is inconsistent.
Finally, we have $F(T, x_1 \wedge x_2) \equiv y_1 \wedge y_2$, $F(T, x_1 \wedge \overline{x_2}) \equiv \top$, $F(T, \overline{x_1} \wedge x_2) \equiv \overline{y_2}$,
and $F(T, \overline{x_1} \wedge \overline{x_2}) \equiv \top$. Thus, $\Sigma$ is fact-compliant with $T$ on every instance, but $(0, 1)$.
\end{example}

With these definitions in hand, the notion of rectification operator can be defined as follows:

\begin{definition}[\textbf{rectification operator}]\label{def:rectification}
A {\em rectification operator} $\star$ is a mapping associating with two given circuits $T$ and $\Sigma$ from $\L$, where $\Sigma$ has the $XY$-classification property, a circuit from $\L$, noted $\Sigma \star T$ and called a {\em rectified circuit}, such that:
	\begin{description}
		\item[\textbf{(RE1)}] $\Sigma \star T$ has the $XY$-classification property;
		\item[\textbf{(RE2)}] If $\Sigma$ is fact-compliant with $T$ on $\vec x \in \vec X$,\\ then $(\Sigma \star T)(\vec x) \equiv \Sigma(\vec x)$;
		\item[\textbf{(RE3)}]	
		For any $\vec x \in \vec X$, $(\Sigma \star T)(\vec x) \models F(T, \vec x)$;
		\item[\textbf{(RE4)}] If $T$ is inconsistent, then $\Sigma \star T \equiv \Sigma$;
		\item[\textbf{(RE5)}] If $\Sigma \equiv \Sigma'$ and $T \equiv T'$, then $\Sigma \star T \equiv \Sigma' \star T'$;
		\item[\textbf{(RE6)}] $\Sigma \star T \equiv (\exists \overline{X \cup Y}. \Sigma) \star (\exists \overline{X \cup Y}. T)$.
	\end{description}
\end{definition}

The rationale for those postulates is presented in \cite{DBLP:conf/ijcai/Coste-MarquisM21}.
Roughly, \textbf{(RE1)} is a closure condition: it asks that any rectified classification circuit is still a classification circuit.
\textbf{(RE2)} is a minimal change condition, stating that the classification of any $\vec x$ as achieved by $\Sigma$ 
should not be modified by the rectification operation whenever $\Sigma$ is fact-compliant with $T$ on $\vec x$.  
\textbf{(RE3)} is a success condition: it demands that the rectified circuit $\Sigma \star T$ is fact-compliant with $T$ on every $\vec x$.
\textbf{(RE4)} is a non-triviality condition; it deals with the case when $T$ is inconsistent; in such a situation, a minimal change of $\Sigma$ consists in not modifying it at all.
\textbf{(RE5)} is a standard principle of irrelevance of syntax.
Finally, \textbf{(RE6)} is a relevance condition: it states that the result of rectifying $\Sigma$ by $T$ must not depend on the variables outside $X \cup Y$.

\begin{example}
Let us consider again the classification circuit $\Sigma$ and the formula $T$ presented at Example \ref{ex:example1}.
Since $\Sigma$ is fact-compliant with $T$ on every instance, but $(0,1)$, \textbf{(RE2)} imposes that $(\Sigma \star T)((1,1)) \equiv y_1 \wedge y_2$,
$(\Sigma \star T)((1,0)) \equiv y_1 \wedge \overline{y_2}$, and $(\Sigma \star T)((0,0)) \equiv \overline{y_1} \wedge \overline{y_2}$.
\textbf{(RE3)} requires that $(\Sigma \star T)((0,1)) \models \overline{y_2}$. Finally, \textbf{(RE1)} ensures that $(\Sigma \star T)((0,1)) \equiv \overline{y_1} \wedge \overline{y_2}$ or $(\Sigma \star T)((0,1)) \equiv y_1 \wedge \overline{y_2}$. Accordingly, the classification $\overline{y_1} \wedge y_2$ of the instance $(0,1)$ as achieved by $\Sigma$ can be rectified in \emph{two} distinct ways in order to enforce that $\overline{y_2}$ holds. Indeed, since no independence assumptions are made about the labels of $Y$,  it can make sense to change the truth value of label $y_1$ when changing the truth value of label $y_2$. The situation is similar to what happens in belief revision, where revising $\overline{y_1} \wedge y_2$ by $\overline{y_2}$ may lead to $y_1 \wedge  \overline{y_2}$ without questioning the satisfaction of the 
revision postulates.
\end{example}

\section{A Characterization Theorem}\label{sec:carac}

Unlike what happens in the general case (as exemplified above), there is a \emph{unique operator $\star$} satisfying 
the rectification postulates in the restricted case when $Y$ contains a single label:

\begin{proposition}\label{prop:unique}
If $\L$ is a language of Boolean circuits over a set of propositional variables $\mathit{PS} = X \cup \{y\}$, then there is a unique rectification operator $\star$.
\end{proposition}

\begin{proof}
First of all, because of postulate \textbf{(RE5)}, we know that the syntactic representations of $\Sigma$ and $T$ does not play any role in the
definition of $\Sigma \star T$ ($\star$ is syntax-independent).
Now, when $Y = \{y\}$ is a singleton, we can mainly get rid of $y$ in the representation of the classifier $\Sigma$ and consider it as implicit (this is
usually done in binary classifiers for the sake of simplicity). Indeed, $\Sigma$ is a classification circuit if and only if there exists a circuit
$\Sigma_X$ from $\L_X$ such that $\Sigma \equiv \Sigma_X \Leftrightarrow y$. 
The models of $\Sigma_X$ are precisely those truth assignments $\vec x$ over $X$ such that $\Sigma(\vec x) \equiv y$.
Because of postulate \textbf{(RE1)}, defining $\Sigma \star T$ 
just amounts to pointing out a circuit $\Sigma_X^T$ from $\L_X$, so that $\Sigma \star T \equiv \Sigma_X^T \Leftrightarrow y$.
We now show that, given $\Sigma$ and $T$, the rectification postulates ensure the existence of a unique circuit $\Sigma_X^T$ up to logical equivalence.
Let $\vec x \in \vec X$. Since $T$ is a circuit from $\L$ and $Y = \{y\}$, $T(\vec x)$ is equivalent to $y$, $\overline{y}$, $\top$, or $\bot$.  
Accordingly,  $F(T, \vec x)$ is equivalent to $T(\vec x) \equiv y$, $T(\vec x) \equiv \overline{y}$, or to $\top$, so that $F(T, \vec x)$ is equivalent 
to $\top$ precisely when it is not equivalent to $T(\vec x)$.
Because of postulate \textbf{(RE1)}, $(\Sigma \star T)(\vec x)$ is equivalent to $y$ or to $\overline{y}$.
By definition, $\Sigma$ is fact-compliant with $T$ on $\vec x$ precisely when $F(T, \vec x)$ is equivalent to $\top$ or $\Sigma(\vec x)$ is equivalent
to $T(\vec x)$, and in this case, because of \textbf{(RE2)}, one must have $(\Sigma \star T)(\vec x) \equiv \Sigma(\vec x)$.
Thus, for any $\vec x \in \vec X$ such that $F(T, \vec x)$ is equivalent to $\top$ or $\Sigma(\vec x)$ is equivalent
to $T(\vec x)$, $\vec x$ is a model of $\Sigma_X^T$ if and only if $\vec x$ is a model of $\Sigma_X$. 
The remaining case, i.e., when $F(T, \vec x)$ is not equivalent to $\top$ and $\Sigma(\vec x)$ is not equivalent
to $T(\vec x)$, can be simplified as $\Sigma(\vec x)$ is not equivalent to $F(T, \vec x) \equiv T(\vec x)$. 
Because of postulate \textbf{(RE4)}, in this case, the class of $\vec x$ must be switched (from positive to negative, or
vice-versa), so that $\vec x$ is a model of $\Sigma_X^T$ if and only if $\vec x$ is not a model of $\Sigma_X$.
This shows that $\Sigma_X^T$ is unique up to logical equivalence, or equivalently that there exists a unique
rectification operator $\star$. Note that $\star$ trivially satisfies \textbf{(RE4)} since if $T$ is inconsistent, 
$F(T, \vec x)$ is equivalent to $\top$ for every $\vec x \in \vec X$, and $\star$ trivially satisfies \textbf{(RE6)}
since $\L$ is built solely upon variables from $X$ and $Y$ (thus, $\exists \overline{X \cup Y}. \Sigma$ is equivalent to
$\Sigma$ and $\exists \overline{X \cup Y}. T$ is equivalent to $T$).
\end{proof}

\begin{example}\label{ex:example2}
As a matter of illustration, let us consider again the loan allocation scenario with Alice and Bob, as sketched in the introduction.
Let us suppose that the predictor $f$ furnished by the bank labels an instance positive when it corresponds to a customer who has high incomes ($\overline{x_1}$) but has not reimbursed a previous loan ($\overline{x_2}$), or (which looks more risky) a customer who has low incomes ($x_1$) and has some debts ($x_3$).
%\begin{itemize}
%\item $x_1 = 1$: has low income
%\item $x_2 = 1$: has reimbursed a previous loan
%\item $x_3 = 1$: has debts
%\item $y = 1$: to grant the loan
%\end{itemize}
Suppose also that Alice’s expertise consists of two decision rules stating that if a customer has low incomes but no debts, the loan can be granted, while if a customer has not reimbursed a previous loan, the loan should not be granted.

%\item $\Sigma(y) = (\overline{x_1} \wedge \overline{x_2}) \vee (x_1 \wedge x_3)$
%\item $T = ((x_1 \wedge \overline{x_3}) \Rightarrow y) \wedge (\overline{x_2} \Rightarrow \overline{y}) $
%\item $T(y) \equiv x_2$
%\item $T(\overline{y})  \equiv \overline{x_1} \vee x_3$

Formally, the predictor $f$ can be represented by the classification circuit $\Sigma = \Sigma_X \Leftrightarrow y$ where $\Sigma_X = (\overline{x_1} \wedge \overline{x_2}) \vee (x_1 \wedge x_3)$. Alice's expertise can be represented by the formula $T = ((x_1 \wedge \overline{x_3}) \Rightarrow y) \wedge (\overline{x_2} \Rightarrow \overline{y})$ encoding her two decision rules.
%Let $T = (\overline{x_1} \wedge \overline{y}) \vee (x_2 \wedge y) \vee (x_3 \wedge \overline{y})$. 
For every instance $\vec x \in \vec X$, Table \ref{tab:example}  indicates from left to right, whether or not $\Sigma$ classifies $\vec x$ as positive (this is the case precisely 
when $\Sigma(\vec x) \equiv y$), the constraint imposed by $T$ on the way $\vec x$ should be classified (i.e., as positive when $T(\vec x) \equiv y$ and as negative when $T(\vec x) \equiv \overline{y}$), the facts $F(T, \vec x)$ that hold in $T$ under $\vec x$, and finally whether or not $\Sigma \star T$ classifies $\vec x$ as positive (this is the case precisely when $(\Sigma \star T)(\vec x) \equiv y$).

\begin{table}
\begin{center}
\scalebox{0.8}{
\begin{tabular}{ccccc}
%\hline
%~ & & & & \\
$\vec x$ & $\Sigma(\vec x)$ & $T(\vec x)$ & $F(T, \vec x)$ & $(\Sigma \star T)(\vec x)$\\
%~ & & & & \\
%\hline
%~ & & & & \\
\midrule
$000$ & $y$ & $\overline{y}$ & $\overline{y}$ & $\overline{y}$\\
$001$ & $y$ & $\overline{y}$ & $\overline{y}$ & $\overline{y}$\\
$010$ & $\overline{y}$ & $\top$ & $\top$ & $\overline{y}$\\
$011$ & $\overline{y}$ & $\top$ & $\top$ & $\overline{y}$\\
$100$ & $\overline{y}$ & $\bot$ & $\top$ & $\overline{y}$\\
$101$ & $y$ & $\overline{y}$ & $\overline{y}$ & $\overline{y}$\\
$\textcolor{orange}{110}$ & $\textcolor{orange}{\overline{y}}$ & $\textcolor{orange}{y}$ & $\textcolor{orange}{y}$ & $\textcolor{orange}{y}$\\
$111$ & $y$ & $\top$ & $\top$ & $y$\\
\midrule
%~ & & & & \\
%\hline
\end{tabular}
}
\end{center}
\caption{The credit scoring scenario, with Alice and Bob.}\label{tab:example}
\end{table}

The instance $\vec x = (x_1 = 1, x_2 = 1, x_3 = 0)$ in orange in the table corresponds to Bob. The classification circuit considered at start classifies $\vec x$ as a negative instance ($\Sigma(\vec x) \equiv \overline{y}$). Contrastingly, the rectified classification circuit $\Sigma \star T$ once $T$ has been taken into account classifies $\vec x$ as positive ($(\Sigma \star T)(\vec x) \equiv y$), as it is expected. One can observe by looking at the table that no specific assumptions are made about what $T$ must say about $y$ under a partial assignment $\vec x$. Thus, the information conveyed by $T$ about $\vec x$ can be trivial, i.e., equivalent to $\top$ (this is the case for instance for $\vec x = (x_1 = 0, x_2 = 1, x_3 = 0)$) or contradictory - equivalent to $\bot$ - (this is the case for $\vec x = (x_1 = 1, x_2 = 0, x_3 = 0)$ since Alice's two decision rules are triggered under this assignment, and those rules have conflicting conclusions).
\end{example}

Note that when $Y$ is a singleton, for every $\vec x \in \vec X$ such that $T(\vec x)$ is consistent, we have
$T(\vec x) \equiv F(T, \vec x)$. Then \textbf{(RE3)} shows immediately that for every $\vec x \in \vec X$ such that $T(\vec x)$ is consistent,
$\Sigma \star T$ is \emph{knowledge-compliant} with $T$ on $\vec x$, i.e., $(\Sigma \star T)(\vec x) \models T(\vec x)$ \cite{DBLP:conf/ijcai/Coste-MarquisM21}.

\section{Representing Rectified Classifiers}\label{sec:representing}

Some rectification operators have been put forward in \cite{DBLP:conf/ijcai/Coste-MarquisM21}. Among them is the following $\star_D$ operator:

\begin{definition}[$\star_D$]~\label{def:rectif}
Let $\circ_D$ denote Dalal revision operator \cite{Dalal88}.\footnote{Given two propositional representations $\varphi$ and $\alpha$, the models of $\varphi \circ_D \alpha$ consist of the models of $\alpha$ which are as close as possible to $\varphi$ w.r.t. Hamming distance.}
Let $\star_D$ be the mapping associating with $T \in \L$ and a classification circuit $\Sigma \in \L$, a classification circuit $\Sigma \star_D T \in \L$ such that 
$$\Sigma \star_D T \equiv \bigvee_{\vec x \in \vec X} \vec x \wedge (\Sigma \star_D T)(\vec x)$$ where for any $\vec x \in \vec X$, 
$(\Sigma \star_D T)(\vec x) = \Sigma(\vec x) \circ_D F(T, \vec x)$.
\label{def:Dalal-rectif}
\end{definition}

It was already observed that $\star_D$ coincides with other rectification operators pointed out in \cite{DBLP:conf/ijcai/Coste-MarquisM21} 
when $Y$ is a singleton. Thanks to Proposition \ref{prop:unique},
we now know more: there is no rectification operator $\star$ that would be different of $\star_D$. Accordingly, the definition of $\star_D$
induces in a straightforward way a characterization result for the class of rectification operators when $|Y| = 1$.

However, the definition of $\star_D$ above is \emph{not convenient at all from a representation perspective} since the representation $\bigvee_{\vec x \in \vec X} \vec x \wedge (\Sigma \star_D T)(\vec x)$ of the rectified classifier $\Sigma \star_D T$ is of size exponential in $|X|$.
In the following, we explain how a much more compact representation of $\Sigma \star T$ can be derived. 
This representation can be computed in \emph{time linear} in 
$|\Sigma|$ and of $|T|$, and its size also is linear in the size of $|\Sigma|$ and of $|T|$.
Remember that when $Y = \{y\}$, because of \textbf{(RE1)}, one knows that 
there exists a circuit $\Sigma_X^T$ from $\L_X$ so that $\Sigma \star T \equiv \Sigma_X^T \Leftrightarrow y$.
Thus, generating a circuit representing $\Sigma \star T$ boils down to generating a circuit representing $\Sigma_X^T$. 

To do so, one first need to make precise the instances that are classified by $T$ as positive, and those that are classified by $T$ as negative.

\begin{proposition}\label{prop:classifies}
Let $\vec x \in \vec X$ and $T \in \L$. $T$ classifies $\vec x$ as
\begin{itemize}
\item a positive instance if $\vec x \models T(y) \wedge \neg T(\overline{y})$;
\item a negative instance if $\vec x \models T(\overline{y}) \wedge \neg T(y)$.
\end{itemize}
\end{proposition}

\begin{proof}
Let us consider the case of positive instances (the other case is similar). By definition, $T$ classifies $\vec x$ as a positive instance
if and only if $T(\vec x) \equiv y$. This means precisely that the assignment $\omega_{\vec x, y}$ that coincides with $\vec x$ over $X$ 
and sets $y$ to true is a model of $T$ and that the assignment $\omega_{\vec x, \overline{y}}$ that coincides with $\vec x$ over $X$ 
and sets $y$ to false is not a model of $T$ (if both  $\omega_{\vec x, y}$ and $\omega_{\vec x, \overline{y}}$ were models of $T$, then
we would have $T(\vec x) \equiv \top$, and if none of $\omega_{\vec x, y}$ and $\omega_{\vec x, \overline{y}}$ were models of $T$, then
we would have $T(\vec x) \equiv \bot$). But $\omega_{\vec x, y} \models T$ precisely means that $\vec x \models \exists \{y\} . (T \wedge y)$, or equivalently
that $\vec x \models T(y)$. And similarly, $\omega_{\vec x, \overline{y}} \not \models T$ precisely means that 
$\vec x \not \models \exists \{y\} . (T \wedge \overline{y})$, or equivalently that $\vec x \not \models T(\overline{y})$.
Finally, since $T(\overline{y})$ is a circuit from $\L_X$ and $\vec x$ is an assignment over $X$, we have $\vec x \not \models T(\overline{y})$
if and only if $\vec x \models \neg T(\overline{y})$. This concludes the proof.
\end{proof}

On this basis, the following representation of $\Sigma_X^T$ can be derived:

\begin{proposition}\label{prop:representation}
Let $\Sigma_X \in \L_X$ and $T \in \L$. We have 
$$\Sigma_X^T \equiv (\Sigma_X \wedge \neg (T(\overline{y}) \wedge \neg T(y))) \vee (T(y) \wedge \neg T(\overline{y})).$$
\end{proposition}

\begin{proof}
The result comes directly from the identification of the only two reasons according to which an instance $\vec x \in \vec X$ must be
classified as positive by the rectified classifier (i.e., it must be a model of $\Sigma_X^T$):
\begin{itemize}
\item Because of \textbf{(RE2)}, $\vec x$ is a model of $\Sigma_X^T$ when $\vec x$ is a model of $\Sigma_X$ and the 
change formula $T$ does not classify $\vec x$ as negative (hence, $\Sigma_X \Leftrightarrow y$ is fact-compliant with $T$ on $\vec x$).
By construction, given Proposition \ref{prop:classifies}, every such model is a model of $\Sigma_X \wedge \neg (T(\overline{y}) \wedge \neg T(y))$.
\item Because of \textbf{(RE3)}, $\vec x$ is a model of $\Sigma_X^T$ when $T$ classifies $\vec x$ as a positive instance.
By construction, given Proposition \ref{prop:classifies}, every such model is a model of $T(y) \wedge \neg T(\overline{y})$.
\end{itemize}
\end{proof}

The rationale of this characterization of $\Sigma_X^T$ is as follows. For an instance $\vec x$ to be classified as positive by the rectified classification circuit,
it must be the case that either $T$ consistently asks for it (this corresponds to the disjunct $T(y) \wedge \neg T(\overline{y})$), or that the classification
circuit considered at start classifies $\vec x$ as positive, provided that $T$ does not consistently ask $\vec x$ to be classified as negative (this corresponds to the
disjunct $\Sigma_X \wedge \neg (T(\overline{y}) \wedge \neg T(y))$). Such a construction is reminiscent to the representation of STRIPS-like actions using propositional formulae, thus asking to make precise each situation where a fluent holds so as to handle the frame problem.

From Proposition \ref{prop:representation}, since the conditioning transformation on circuits can be achieved in linear time,
$\Sigma_X^T \Leftrightarrow y$ (where $\Sigma_X^T$ is provided by Proposition \ref{prop:representation}) is a circuit
of $\L$ equivalent to $\Sigma \star T$ and computable in time linear in $|\Sigma|+|T|$. Its size is also linear in $|\Sigma|+|T|$, as expected.

\begin{example}
Let us consider $\Sigma$ and $T$ as in Example \ref{ex:example2}. We have $T(y) \equiv x_2$ and $T(\overline{y}) \equiv \overline{x_1} \vee x_3$.
Thus, we get 
$$\Sigma_X^T \equiv (\underbrace{(\overline{x_1} \wedge \overline{x_2}) \vee (x_1 \wedge x_3)}_{\Sigma_X} 
\wedge \neg (\underbrace{(\overline{x_1} \vee x_3)}_{T(\overline{y})} \wedge \neg \underbrace{x_2}_{T(y)}))$$
$$ \vee (\underbrace{x_2}_{T(y)} \wedge \neg(\underbrace{(\overline{x_1} \vee x_3)}_{T(\overline{y})}).$$
This circuit can be simplified as $\Sigma_X^T \equiv x_1 \wedge x_2$.
One can check in Table \ref{tab:example} (rightmost colum) that the models of $\Sigma_X^T$ are precisely those 
$\vec x \in \vec X$ such that $(\Sigma \star T)(\vec x) \equiv y$. Stated otherwise, for the rectified classification circuit
$\Sigma \star T$, the clients for which a loan can be granted are those having low incomes provided
that they have reimbursed a previous loan.
\end{example}

Note that if $\Sigma_X$ and $T$ are formulae (and not Boolean circuits) in Proposition \ref{prop:representation}, then 
the resulting characterization of $\Sigma_X^T$ also is a formula (indeed, conditioning a formula leads to a formula).
Furthermore, whenever $\Sigma_X$ and $T$ belongs to a
class $\mathcal{C}$ of circuits that offers in polynomial time the transformations of negation ($\nc$), bounded conjunction ($\bac$),
and bounded disjunction ($\boc$) \cite{DarwicheMarquis02}, a representation of $\Sigma_X^T$ in $\mathcal{C}$ 
can be \emph{derived in polynomial time} from $\Sigma_X$ and $T$. 

Notably, focusing on a restricted class of circuits $\mathcal{C}$ is not mandatory
for ensuring tractable classification: when $\Sigma_X^T$ is in $\L_X$, deciding whether or not
$\vec x \in \vec X$ is classified as positive by $\Sigma \star T$ amounts to testing whether or not $\vec x$ is a
model of $\Sigma_X^T$, and such a model checking test can be done in time linear in the size of the input.

However, considering specific classes of circuits can prove valuable for other reasons, especially from an eXplainable
AI perspective (see e.g., \cite{Audemardetal20,DBLP:conf/nips/BarceloM0S20,DBLP:conf/aaai/ArenasBBM21,DBLP:conf/aaai/BroeckLSS21,DBLP:conf/kr/HuangII021}). 
Among the classes of circuits offering  $\nc$, $\bac$, and $\boc$ are \sdd, the class
of sentential decision diagrams \cite{Darwiche11}, \obdd, the class
of ordered binary decision diagrams \cite{Bryant86}, but also \dt, the class of decision trees, and more generally \adt, the class of
affine decision trees \cite{Koricheetal13}. 
Any Boolean circuit can be represented in \sdd, \obdd, \adt\ and  \dt. Thus, considering those languages  for representing 
the change formula $T$ that triggers the rectification operation allows us to accept as input any possible $T$ (up to logical equivalence). 
Of course, it is not the case that every Boolean circuit $T$
has a representation in \sdd, \obdd, \adt\ or \dt\ that is of size polynomial in $|T|$, but ``simple'' change formulae $T$ (e.g., clauses or terms)
can be turned in linear time into equivalent representations in \sdd, \obdd, \adt, and \dt. 
For instance, a classification rule like $(x_1 \wedge \overline{x_3}) \Rightarrow y$ 
(equivalent to the clause $\overline{x_1} \vee x_3 \vee y$) that is entailed by the formula $T$ considered in Example \ref{ex:example2} 
could be easily handled.

The case of \dt\ is of particular interest since it corresponds 
to a well-known ML model \cite{DBLP:books/wa/BreimanFOS84,DBLP:journals/ml/Quinlan86}, that also serves as a key component 
of other ML models, especially random forests \rf\ \cite{Breiman01} and boosted trees \cite{DBLP:journals/jcss/FreundS97}.
Furthermore, \dt\ is a much more intelligible model than most of 
Boolean classifiers \cite{DBLP:conf/kr/AudemardBBKLM21}, and \rf\ also offers some tractable 
explanation facilities (via the notion of majoritary reason) \cite{Audemardetal22}.
Given the significance of those ML models, the existence of polynomial-time algorithms for rectifying 
decision trees and random forests\footnote{Rectifying a random forest or an (Adaboost-style) boosted tree simply amounts to rectifying every decision tree in it.} is 
a \emph{noteworthy consequence} of Proposition \ref{prop:representation}.

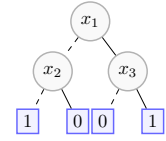
\begin{figure} %[t!]
\centering 
%\begin{subfigure}[b]{0.25\textwidth}%,valign=t}
%    \centering
    \scalebox{0.7}{
          \begin{tikzpicture}[scale=0.475, roundnode/.style={circle, draw=gray!60, fill=gray!5, thick, minimum size=5mm},
          squarednode/.style={rectangle, draw=blue!60, fill=blue!5, thick, minimum size=3mm}]
            \node[roundnode](root) at (2.5,7){$x_1$};
%            \node at (0.5,7){$\Sigma_X$};
            \node[roundnode](n1) at (1,5){$x_2$};
            \node[roundnode](n2) at (4,5){$x_3$};
            \node[squarednode](n11) at (0,3){$1$};
            \node[squarednode](n12) at (2,3){$0$};
            \node[squarednode](n21) at (3,3){$0$};
            \node[squarednode](n22) at (5,3){$1$};
            \draw[dashed] (root) -- (n1);
            \draw(root) -- (n2);
            \draw[dashed] (n1) -- (n11);
            \draw(n1) -- (n12); 
            \draw[dashed] (n2) -- (n21);
            \draw(n2) -- (n22);    
         \end{tikzpicture}
     }
        \caption{A decision tree representing $\Sigma_X$.}\label{fig:figure-a}
        %\caption{}\label{fig:figure-a}
%\end{subfigure}
\end{figure}

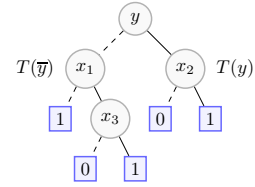
\begin{figure} %[t!]
\centering 
%\begin{subfigure}[b]{0.25\textwidth}
%    \centering
    \scalebox{0.7}{
          \begin{tikzpicture}[scale=0.475, roundnode/.style={circle, draw=gray!60, fill=gray!5, thick, minimum size=5mm},
          squarednode/.style={rectangle, draw=blue!60, fill=blue!5, thick, minimum size=3mm}]                  
            \node[roundnode](root2) at (14,7){$y$};
%            \node at (11.5,7){$T$};
            \node[roundnode](n2-1) at (12,5){$x_1$};  
            \node at (10,5){$T(\overline{y})$};
            \node[roundnode](n2-2) at (16,5){$x_2$};
            \node at (18,5){$T(y)$};
            \node[squarednode](n2-11) at (11,3){$1$};
            \node[roundnode](n2-12) at (13,3){$x_3$}; 
            \node[squarednode](n2-21) at (15,3){$0$};
            \node[squarednode](n2-22) at (17,3){$1$};
            \node[squarednode](n2-121) at (12,1){$0$};
            \node[squarednode](n2-122) at (14,1){$1$};
            \draw[dashed] (root2) -- (n2-1);
            \draw(root2) -- (n2-2);
            \draw[dashed] (n2-1) -- (n2-11);
            \draw(n2-1) -- (n2-12);    
            \draw[dashed] (n2-2) -- (n2-21);
            \draw(n2-2) -- (n2-22); 
            \draw[dashed] (n2-12) -- (n2-121);
            \draw(n2-12) -- (n2-122);          
            \end{tikzpicture}
     }
        \caption{A decision tree representing $T$ (right). The subtrees of $T$ rooted at nodes $T(\overline{y})$ and $T(y)$, respectively, are decision trees representing $T(\overline{y})$ and $T(y)$.}\label{fig:figure-b}
        %\caption{}\label{fig:figure-b}
%\end{subfigure}
\end{figure}

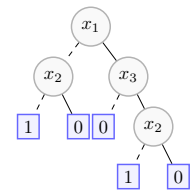
\begin{figure} %[t!]
\centering 
%\begin{subfigure}[b]{0.25\textwidth}
%    \centering
    \scalebox{0.7}{
          \begin{tikzpicture}[scale=0.475, roundnode/.style={circle, draw=gray!60, fill=gray!5, thick, minimum size=5mm},
          squarednode/.style={rectangle, draw=blue!60, fill=blue!5, thick, minimum size=3mm}]           
            \node[roundnode](root) at (2.5,7){$x_1$};
 %           \node at (-2,7){$T(y) \wedge \neg T(\overline{y})$};
            \node[roundnode](n1) at (1,5){$x_2$};
            \node[roundnode](n2) at (4,5){$x_3$};
            \node[squarednode](n11) at (0,3){$1$};
            \node[squarednode](n12) at (2,3){$0$};
            \node[squarednode](n21) at (3,3){$0$};
            \node[roundnode](n22) at (5,3){$x_2$};
            \node[squarednode](n221) at (4,1){$1$};
            \node[squarednode](n222) at (6,1){$0$};           
            \draw[dashed] (root) -- (n1);
            \draw(root) -- (n2);
            \draw[dashed] (n1) -- (n11);
            \draw(n1) -- (n12); 
            \draw[dashed] (n2) -- (n21);
            \draw(n2) -- (n22);    
            \draw[dashed] (n22) -- (n221);
            \draw(n22) -- (n222); 
            \end{tikzpicture}
     }   
        \caption{A decision tree representing $T(\overline{y}) \wedge \neg T(y)$.}\label{fig:figure-c}
        %\caption{}\label{fig:figure-c}
%\end{subfigure} 
\end{figure}

\begin{figure} %[t!]
\centering 
%\begin{subfigure}[b]{0.25\textwidth}
%    \centering
    \scalebox{0.7}{
          \begin{tikzpicture}[scale=0.475, roundnode/.style={circle, draw=gray!60, fill=gray!5, thick, minimum size=5mm},
          squarednode/.style={rectangle, draw=blue!60, fill=blue!5, thick, minimum size=3mm}]                       
            \node[roundnode](root2) at (11,7){$x_2$};
%            \node at (15,7){$T(\overline{y}) \wedge \neg T(y)$};
            \node[squarednode](n2-1) at (10,5){$0$};
            \node[roundnode](n2-2) at (12,5){$x_1$};
            \node[squarednode](n2-21) at (11,3){$0$};
            \node[roundnode](n2-22) at (13,3){$x_3$};
            \node[squarednode](n2-221) at (12,1){$1$};
            \node[squarednode](n2-222) at (14,1){$0$};           
            \draw[dashed] (root2) -- (n2-1);
            \draw(root2) -- (n2-2);
            \draw[dashed] (n2-2) -- (n2-21);
            \draw(n2-2) -- (n2-22); 
            \draw[dashed] (n2-22) -- (n2-221);
            \draw(n2-22) -- (n2-222);          
           \end{tikzpicture}
     }
        \caption{A decision tree representing $T(y) \wedge \neg T(\overline{y})$.}\label{fig:figure-d}
        %\caption{}\label{fig:figure-d}
%\end{subfigure} 
\end{figure}
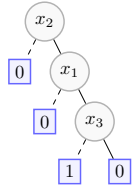

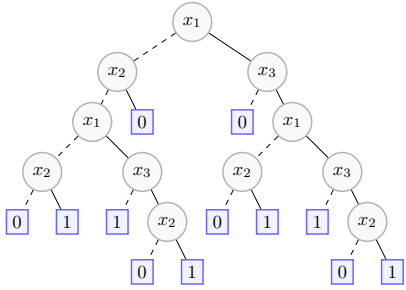
\begin{figure} %[t!]
\centering 
%\begin{subfigure}[b]{0.25\textwidth}
%    \centering
    \scalebox{0.7}{
          \begin{tikzpicture}[scale=0.475, roundnode/.style={circle, draw=gray!60, fill=gray!5, thick, minimum size=5mm},
          squarednode/.style={rectangle, draw=blue!60, fill=blue!5, thick, minimum size=3mm}]           
            \node[roundnode](root) at (9,11){$x_1$};
            \node[roundnode](n1) at (6,9){$x_2$};
            \node[roundnode](n2) at (12,9){$x_3$};
            \node[roundnode](n11) at (5,7){$x_1$};
            \node[squarednode](n12) at (7,7){$0$};
            \node[roundnode](n111) at (3,5){$x_2$};
            \node[roundnode](n112) at (7,5){$x_3$};                       
            \node[squarednode](n1111) at (2,3){$0$};
            \node[squarednode](n1112) at (4,3){$1$};
            \node[squarednode](n1121) at (6,3){$1$};
            \node[roundnode](n1122) at (8,3){$x_2$};
            \node[squarednode](n11221) at (7,1){$0$};
            \node[squarednode](n11222) at (9,1){$1$};          
            
            \node[squarednode](n21) at (11,7){$0$};
            \node[roundnode](n22) at (13,7){$x_1$};
            \node[roundnode](n221) at (11,5){$x_2$};
            \node[roundnode](n222) at (15,5){$x_3$};                       
            \node[squarednode](n2211) at (10,3){$0$};
            \node[squarednode](n2212) at (12,3){$1$};
            \node[squarednode](n2221) at (14,3){$1$};
            \node[roundnode](n2222) at (16,3){$x_2$};
            \node[squarednode](n22221) at (15,1){$0$};
            \node[squarednode](n22222) at (17,1){$1$};         
                    
            \draw[dashed] (root) -- (n1);
            \draw(root) -- (n2);
            \draw[dashed] (n1) -- (n11);
            \draw(n1) -- (n12); 
            \draw[dashed] (n11) -- (n111);
            \draw(n11) -- (n112);     
            \draw[dashed] (n111) -- (n1111);
            \draw(n111) -- (n1112);              
            \draw[dashed] (n112) -- (n1121);
            \draw(n112) -- (n1122); 
            \draw[dashed] (n1122) -- (n11221);
            \draw(n1122) -- (n11222);       
            
            \draw[dashed] (n2) -- (n21);
            \draw(n2) -- (n22); 
            \draw[dashed] (n22) -- (n221);
            \draw(n22) -- (n222);     
            \draw[dashed] (n221) -- (n2211);
            \draw(n221) -- (n2212);              
            \draw[dashed] (n222) -- (n2221);
            \draw(n222) -- (n2222); 
            \draw[dashed] (n2222) -- (n22221);
            \draw(n2222) -- (n22222);     
           \end{tikzpicture}
     }
        \caption{A decision tree representing $\Sigma_X \wedge \neg(T(\overline{y}) \wedge \neg T(y))$.}\label{fig:figure-e}
        %\caption{}\label{fig:figure-e}
%\end{subfigure}      
\end{figure}

\begin{example}
Considering  Example \ref{ex:example2} again, let us finally illustrate how a decision tree classifier equivalent to 
$\Sigma \star T$ can be generated in polynomial time from $\Sigma$ and $T$. 
Starting with a decision tree $\Sigma$ over $X \cup \{y\}$, a decision tree over $X$ equivalent to 
$\Sigma_X$ (given at Figure \ref{fig:figure-a}) can be obtained by conditioning $\Sigma$ by $y$ since $\Sigma_X \equiv \Sigma(y)$. 
Conditioning a decision tree by a literal $v$ (resp. $\overline{v}$) 
amounts to replacing in 
the tree every decision node over variable $v$ by its right (resp. left) child. Using the conditioning transformation, from the decision tree 
$T$ over $X \cup \{y\}$ at Figure \ref{fig:figure-b}, one can derive efficiently decision trees for  $T(\overline{y})$ and $T(y)$ 
(in Figure \ref{fig:figure-b}, they are 
the subtrees 
rooted at nodes $T(\overline{y})$ and $T(y)$).

On this ground, deriving decision trees equivalent to $T(\overline{y}) \wedge \neg T(y)$ and
$T(y) \wedge \neg T(\overline{y})$, as reported on Figures \ref{fig:figure-c} and \ref{fig:figure-d} (respectively), requires to be
able to negate and to conjoin decision trees. 
Negating a tree consists in replacing each of its $1$-leaves by a $0$-leaf, and vice-versa. Conjoining
two decision trees consists in replacing every $1$-leaf of the first tree by a copy of the second tree. In the general case, the conjunction operation 
may lead to a decision tree that is not simplified, because it is not read-once and may include decision nodes having two identical children \cite{Wegener00}.
However, such a tree can be simplified in linear time into an equivalent tree, using the following rules whenever applicable: on the one hand, every decision node over a variable $x_i$ can be replaced by its left (resp. right) child when it is itself the left (resp. right) child of a decision node over $x_i$ or when it has an ancestor satisfying this property; on the other hand, a decision node having two identical children can be replaced by any of its two children.

Figure \ref{fig:figure-e} presents a decision tree equivalent to $\Sigma_X \wedge \neg(T(\overline{y}) \wedge \neg T(y))$ and obtained by
conjoining the decision tree of $\Sigma_X$ given in Figure \ref{fig:figure-a} (left) with the negation of the decision tree of $T(\overline{y}) \wedge \neg T(y)$
given in Figure \ref{fig:figure-c}. Figure \ref{fig:figure-f} illustrates the effect of the simplification process. 

Figure \ref{fig:figure-g} presents a decision tree equivalent to $(\Sigma_X \wedge \neg(T(\overline{y}) \wedge \neg T(y))) \vee (T(y) \wedge \neg T(\overline{y}))$, obtained by disjoining the decision tree at Figure \ref{fig:figure-f} with the decision tree at Figure \ref{fig:figure-d}. This is achieved by replacing every $0$-leaf of the first tree by a copy of the second tree. The resulting tree is not simplified, and Figure \ref{fig:figure-h} presents an equivalent, yet simplified decision tree (obtained by running the simplification procedure sketched above). By construction, it is equivalent to $\Sigma_X^T$. As expected, one recovers here the condition $x_1 \wedge x_2$ characterizing the positive instances w.r.t. the rectified classification circuit: the clients for which a loan can be granted are those having low incomes provided
that they have reimbursed a previous loan.
Finally, a decision tree equivalent to $\Sigma \star T$ can be generated in linear time
from $\Sigma_X^T$ by replacing every $1$-leaf (resp. $0$-leaf) by a decision node over $y$, with a $0$-leaf (resp. $1$-leaf) as left child and a $1$-leaf (resp. $0$-leaf) as right child.
\end{example}

%\begin{figure*}[t!]
%  \centering
%  \begin{subfigure}[t]{0.5\textwidth}
%    \centering
%    \includegraphics[width=\textwidth]{cachetCactus.pdf}
%    \caption{\label{fig:cactusCachet} \cachet}
%  \end{subfigure}
 
%~\\

\begin{figure} %[t!]
\centering 
%\begin{subfigure}[b]{0.25\textwidth}
%    \centering
    \scalebox{0.7}{
          \begin{tikzpicture}[scale=0.475, roundnode/.style={circle, draw=gray!60, fill=gray!5, thick, minimum size=5mm},
          squarednode/.style={rectangle, draw=blue!60, fill=blue!5, thick, minimum size=3mm}]                  
            \node[roundnode](root2) at (20,11){$x_1$};
            \node[squarednode](n2-1) at (19,9){$0$};
            \node[roundnode](n2-2) at (21,9){$x_3$};
            \node[squarednode](n2-21) at (20,7){$0$};
            \node[roundnode](n2-22) at (22,7){$x_2$};
            \node[squarednode](n2-221) at (21,5){$0$};
            \node[squarednode](n2-222) at (23,5){$1$};           
            \draw[dashed] (root2) -- (n2-1);
            \draw(root2) -- (n2-2);
            \draw[dashed] (n2-2) -- (n2-21);
            \draw(n2-2) -- (n2-22); 
            \draw[dashed] (n2-22) -- (n2-221);
            \draw(n2-22) -- (n2-222);   
           \end{tikzpicture}
     }
        \caption{The decision tree of Figure \ref{fig:figure-e}, once simplified.}\label{fig:figure-f}
        %\caption{}\label{fig:figure-f}
%\end{subfigure}  
\end{figure}
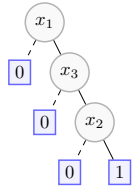   
%~\\

\begin{figure} %[t!]
\centering 
%\begin{subfigure}[b]{0.25\textwidth}
%    \centering
    \scalebox{0.7}{
          \begin{tikzpicture}[scale=0.475, roundnode/.style={circle, draw=gray!60, fill=gray!5, thick, minimum size=5mm},
          squarednode/.style={rectangle, draw=blue!60, fill=blue!5, thick, minimum size=3mm}]       
                                     
            \node[roundnode](root) at (7,9){$x_1$};
            \node[roundnode](n2) at (12,7){$x_3$};
            \node[roundnode](n21) at (8,5){$x_2$};
            \node[roundnode](n22) at (16,5){$x_2$};
            \node[squarednode](n222) at (17,3){$1$};
                 
            \node[roundnode](n1) at (2,7){$x_2$};
            \node[squarednode](n1-1) at (1,5){$0$};
            \node[roundnode](n1-2) at (3,5){$x_1$};
            \node[squarednode](n1-21) at (2,3){$0$};
            \node[roundnode](n1-22) at (4,3){$x_3$};
            \node[squarednode](n1-221) at (3,1){$1$};
            \node[squarednode](n1-222) at (5,1){$0$};        
            
            \node[squarednode](n21-1) at (7,3){$0$};
            \node[roundnode](n21-2) at (9,3){$x_1$};
            \node[squarednode](n21-21) at (8,1){$0$};
            \node[roundnode](n21-22) at (10,1){$x_3$};
            \node[squarednode](n21-221) at (9,-1){$1$};
            \node[squarednode](n21-222) at (11,-1){$0$};         
            
            \node[roundnode](n221) at (15,3){$x_2$};
            \node[squarednode](n221-1) at (14,1){$0$};
            \node[roundnode](n221-2) at (16,1){$x_1$};
            \node[squarednode](n221-21) at (15,-1){$0$};
            \node[roundnode](n221-22) at (17,-1){$x_3$};
            \node[squarednode](n221-221) at (16,-3){$1$};
            \node[squarednode](n221-222) at (18,-3){$0$};                   
            
            \draw[dashed] (root) -- (n1);
            \draw(root) -- (n2);
            \draw[dashed] (n2) -- (n21);
            \draw(n2) -- (n22);
            \draw[dashed] (n22) -- (n221);
            \draw(n22) -- (n222);

            \draw[dashed] (n1) -- (n1-1);
            \draw(n1) -- (n1-2);
            \draw[dashed] (n1-2) -- (n1-21);
            \draw(n1-2) -- (n1-22); 
            \draw[dashed] (n1-22) -- (n1-221);
            \draw(n1-22) -- (n1-222);     
            
            \draw[dashed] (n21) -- (n21-1);
            \draw(n21) -- (n21-2);
            \draw[dashed] (n21-2) -- (n21-21);
            \draw(n21-2) -- (n21-22); 
            \draw[dashed] (n21-22) -- (n21-221);
            \draw(n21-22) -- (n21-222);    
            
            \draw[dashed] (n221) -- (n221-1);
            \draw(n221) -- (n221-2);
            \draw[dashed] (n221-2) -- (n221-21);
            \draw(n221-2) -- (n221-22); 
            \draw[dashed] (n221-22) -- (n221-221);
            \draw(n221-22) -- (n221-222);                
           \end{tikzpicture}
     }
        \caption{A decision tree representing $\Sigma_X^T$.}\label{fig:figure-g}
        %\caption{}\label{fig:figure-g}
%\end{subfigure} 
\end{figure}
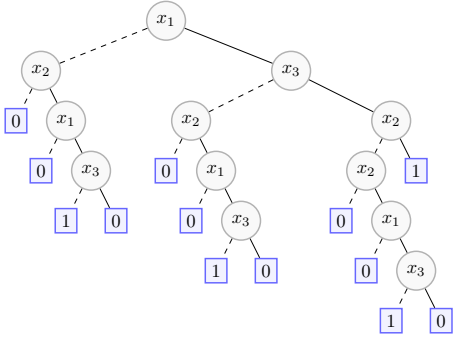

\begin{figure} %[t!]
\centering 
%\begin{subfigure}[b]{0.25\textwidth}
%    \centering
    \scalebox{0.7}{
          \begin{tikzpicture}[scale=0.475, roundnode/.style={circle, draw=gray!60, fill=gray!5, thick, minimum size=5mm},
          squarednode/.style={rectangle, draw=blue!60, fill=blue!5, thick, minimum size=3mm}]       
                                     
            \node[roundnode](root2) at (22,9){$x_1$};
            \node[squarednode](n2-1) at (21,7){$0$};
            \node[roundnode](n2-2) at (23,7){$x_2$};
            \node[squarednode](n2-21) at (22,5){$0$};
            \node[squarednode](n2-22) at (24,5){$1$};       
            \draw[dashed] (root2) -- (n2-1);
            \draw(root2) -- (n2-2);
            \draw[dashed] (n2-2) -- (n2-21);
            \draw(n2-2) -- (n2-22); 
               
           \end{tikzpicture}
     }
        \caption{The decision tree of Figure \ref{fig:figure-g}, once simplified.}\label{fig:figure-h}
        %\caption{}\label{fig:figure-h}
%\end{subfigure}  
          %\caption{\label{fig:figure} Rectifying a decision tree classifier.}
\end{figure}
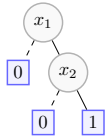

\section{Conclusion}\label{sec:conclusion}

The main contribution of this paper is twofold. On the one hand, we have presented a
characterization theorem for the unique rectification operator $\star$ obtained when dealing with mono-label Boolean classifiers.
On the other hand, we have explained how a classification circuit equivalent to $\Sigma \star T$ can be computed in time linear in the size of $\Sigma$ 
and $T$. Especially, we have also shown that a decision tree equivalent to $\Sigma \star T$ can be computed in time polynomial in the size of $\Sigma$ 
and $T$ when each of $\Sigma$ and $T$ is represented as a decision tree. 

In eXplainable AI, a contrastive explanation for an instance $\vec x$ aims to to explain 
why $\vec x$ has not been classified by the ML model as the explainee expected it (thus, addressing the ``Why not?'' question) 
\cite{Miller19,DBLP:journals/corr/abs-2012-11067}. 
When the explainee is not only surprised by the prediction made by the classifier, but actually believes that the prediction is wrong, pointing out an explanation 
is not enough. A rectification process must take place. We have shown how to achieve this process efficiently when the mono-label Boolean classifier
at hand is a decision tree or a classifier based on such trees (random forests, boosted trees). 
%Given the number of applications involving decision trees and/or random forests, we believe that this result is significant enough to be noticed. 

\medskip

In this work, one started with the basic assumption that the available background knowledge $T$ is more reliable than the classification
circuit $\Sigma$. We believe that it is a reasonable assumption for many scenarios. Especially, 
the assumption is similar to the one considered in AGM belief revision (primacy of the new information). That mentioned, 
just like AGM belief revision is not suited to every revision issue (semi-revision \cite{DBLP:journals/jancl/Hansson97}, promotion \cite{SKM18b}, or improvement \cite{KGP10},  can be used when the assumption does not hold), 
it would be interesting to determine how to relax the basic assumption and deal with pieces of expert knowledge that might be
faulty or conflicting. This is left for further research.

%\newpage

\section*{Acknowledgements}
This work has benefited from the support of the AI Chair EXPE\textcolor{orange}{KC}TA\-TION (ANR-19-CHIA-0005-01) of the French National Research Agency (ANR).
It was also partially supported by TAILOR, a project funded by EU Horizon 2020 research and innovation programme under GA No 952215.

%\newpage
\bibliographystyle{named}
\bibliography{rectification-DT-arXiv}

%\appendix

\end{document}